\documentclass[conference]{IEEEtran}
\IEEEoverridecommandlockouts
\usepackage{cite}
\usepackage{amsmath,amssymb,amsfonts}
\usepackage{algorithmic}
\usepackage{graphicx}
\usepackage{textcomp}
\usepackage{xcolor}
\usepackage{xspace}

\def\BibTeX{{\rm B\kern-.05em{\sc i\kern-.025em b}\kern-.08em
    T\kern-.1667em\lower.7ex\hbox{E}\kern-.125emX}}

\newcommand{\myparagraph}[1]{\vspace{0.15cm}\noindent\textbf{#1.}}

\def\eg{\emph{e.g.\,}}
\def\ie{\emph{i.e.\,}}

\begin{document}

\title{Deep Visual Domain Adaptation}

\author{{Gabriela Csurka }\\
Naver Labs Europe, Meylan, France \\
https://europe.naverlabs.com\\
\emph{gabriela.csurka@naverlabs.com}}

\maketitle

\begin{abstract}
Domain adaptation (DA) aims at improving the performance of a model on target domains by transferring the knowledge contained in different but related source domains. With recent advances in deep learning models which are extremely data hungry, the interest for visual DA  has significantly increased in the last decade and the number of related work in the field exploded. The aim of this paper, therefore,  is to  give a comprehensive  overview of deep domain adaptation methods for computer vision applications. First, we detail and compared different possible ways of exploiting deep architectures for domain adaptation. Then, we propose an overview of recent trends in deep visual DA. Finally, we mention a few improvement strategies, orthogonal to these methods, that can be applied to these models. While we mainly focus on image classification, we give pointers to  papers that extend these ideas for other applications  such as semantic segmentation,  object detection, person re-identifications, and others.

\end{abstract}


\section{Introduction}

\begin{figure}[ttt]
\begin{center}
\includegraphics[width=0.48\textwidth]{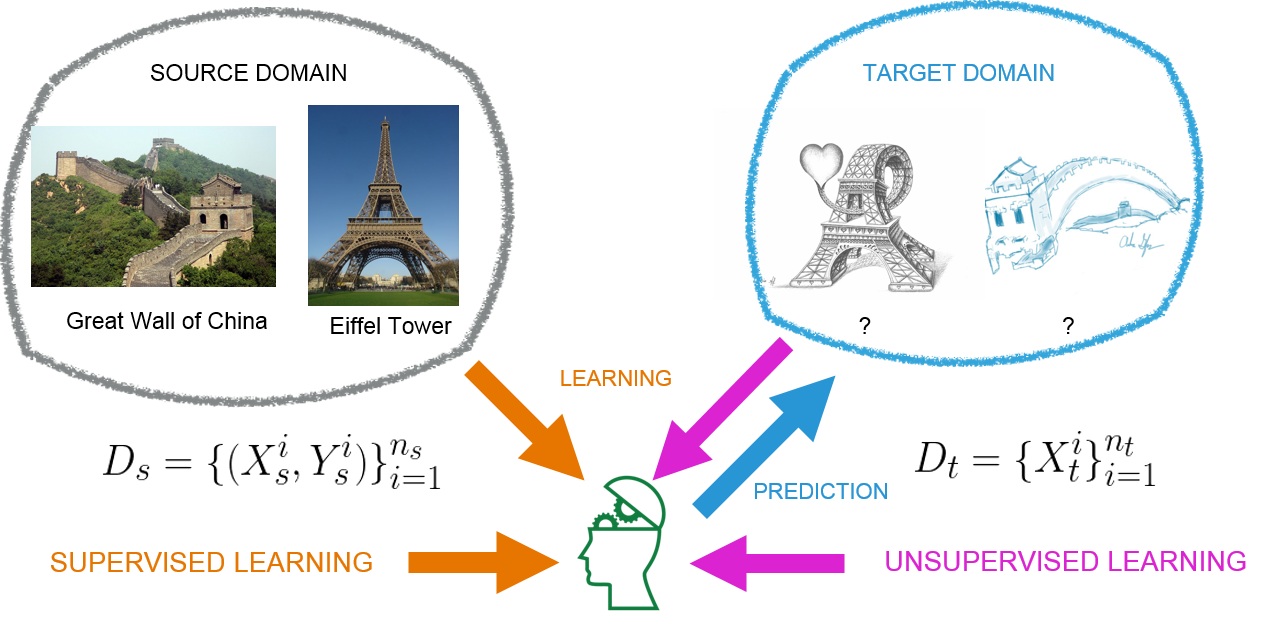}
\caption{Domain adaptation is a machine learning technique where knowledge from a labeled source domain is leveraged  to learn a model for an unlabeled target domain. It is assumed that there is a distribution mismatch between domains but the task (\eg class labels) is shared between domains.}
\vspace{-0.4cm}
\label{fig:USDA}
\end{center}
\end{figure}

While recent advances in deep learning  yielded  a significant boost in performance in most computer  vision tasks, this success  depends a lot on the availability of a large amount of well-annotated
training data.   As the cost of acquiring data labels remains high,  amongst alternative solutions,  domain adaptation approaches have been  proposed, where the main idea is to exploit the unlabeled data within the same domain together with annotated data from a different yet related domain. Yet, because learning from the new domain might suffer from distribution mismatch  between the two domains,  it is necessary to  adapt the model learned on the labelled {\em source}   to the actual {\em target} domain as pictured in Fig. \ref{fig:USDA}. 

With the recent progress on deep learning, a significant performance  boost  over previous state-of-the art
of visual  categorization systems was observed.
In parallel, it was shown that features extracted from
the activation layers of these deep networks can be re-purposed for novel tasks or domains~\cite{DonahueICML14DeCAFActivationFeature} even when the new task/domain differs from the task/domain originally used to train the model. This  is  because
deep neural networks learn more abstract and more robust representations, they encode category level information and remove, to a certain measure,  the domain bias \cite{BengioPAMI13RepresentationLearningReviewNewPerspectives,YosinskiNIPS14HowTransferableDNN}.
Hence, these representations  are  more transferable to new tasks/domains because they disentangle the factors of  variations in underlying data samples while grouping them  hierarchically according to their relatedness with invariant factors.

\begin{figure}[ttt]
\begin{center}
\includegraphics[width=0.5\textwidth]{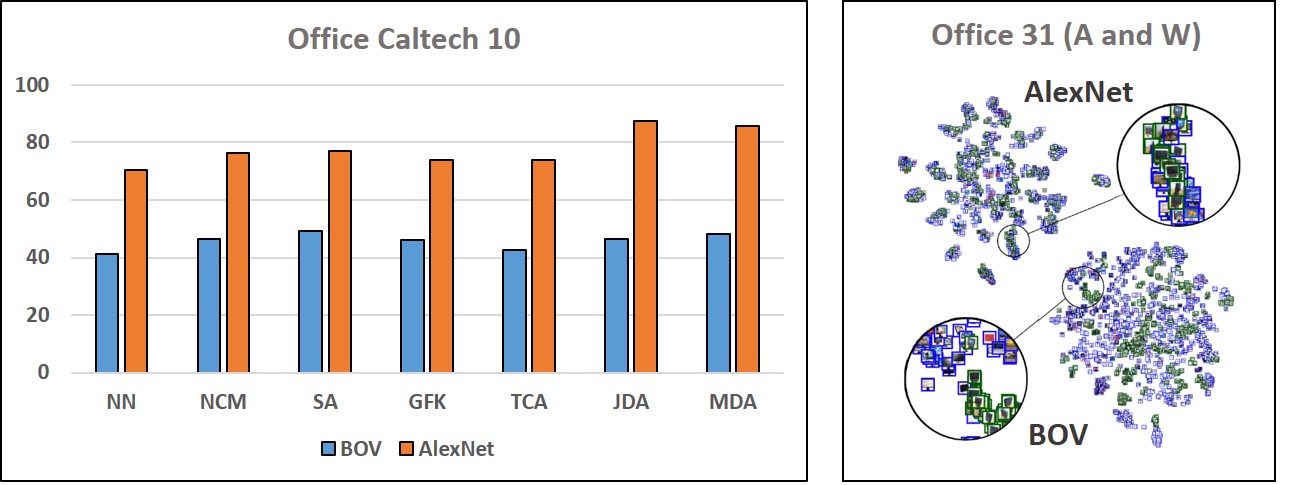}
\caption{Left: Results show that nearest neighbor (NN) classifier results with AlexNet \cite{KrizhevskyNIPS12ImagenetDeepCNN} without any adaptation on the Office+Caltech \cite{GongCVPR12GeodesicFlowKernel} dataset  outperform by a large margin classical shallow DA methods using
the SURF-BOV features originally provided with these datasets. Right:  we show 
Amazon (A)  and Webcam (W) data from  the Office 31 \cite{SaenkoECCV10AdaptingVisualCategoryModels} benchmark set clustered together 
with SURF-BOV and AlexNet  features. We  
can see that  the two domains are  much better clustered  with deep features  then with  SURF-BOV.}
\label{fig:BovAlexNet}
\vspace{-0.4cm}
\end{center}
\end{figure}

These image representations, in general obtained by
training the model in a fully supervised manner on large-scale annotated datasets, in particular 
ImageNet~\cite{RussakovskyIJCV15ImagenetVisualRecognChallenge}, can therefore be directly  used to build stronger baselines for domain adaptation methods.
Indeed, by simply training a linear classifier with such representations obtained from activation layers \cite{DonahueICML14DeCAFActivationFeature},  and  with no further adaptation to the target set,  yields in general  significantly better results than most shallow
DA  models trained with previously used handcrafted, generally bag of visual words (BOV) \cite{CsurkaECCVWS04VisualCategorizationBagsKeypoints},  representations. 
In  Fig. \ref{fig:BovAlexNet} we illustrate this using the AlexNet architecture \cite{KrizhevskyNIPS12ImagenetDeepCNN}, however representations obtained with deeper models \cite{SimonyanX14VeryDeepConvolutionalNetworks,HeCVPR16DeepResidualLearning,SzegedyCVPR15GoingDeeperWithConvolutions} provide even better  performance and generalization capacity~\cite{CsurkaTASKCV17DiscrepancyBasedNetworksUnsupervisedDAComparativeStudy}.

While using directly these models trained on the source provides already relatively good results on the target
datasets, especially  when the domain shift is moderate, for more challenging problems, \eg adaptation between images and paintings, drawings, clip art or
sketches  \cite{CastrejonCVPR16LearningAlignedCrossModal,LiCVPR17DeeperBroaderArtierDG,CsurkaTASKCV17DiscrepancyBasedNetworksUnsupervisedDAComparativeStudy},  a classifier trained even with such deep features would have  difficulties to handle the domain differences. Therefore,  the need for  alternative  solutions that directly handle the domain shift remains the preferred solution.  

Therefore, in which follows we  first  discuss and compare
different strategies about how
to exploit deep architectures for  domain adaptation. Then, we provide an overview of recent trends in deep visual domain adaptation.  Finally, we evoke
a few  strategies,  orthogonal to the deep DA architecture design,  that can be applied to improve those models.

\section{Deep learning strategies}
\label{sec:DeepStrategies}

There are several ways to exploit deep models to handle 
the domain mismatch between the source and the target set,
that can be grouped in four main categories: 1) shallow methods using deep features, 2) using fine-tuned deep architectures, 3) shallow methods using fine-tuned deep features and 4) deep domain adaptation models. 

\myparagraph{Shallow DA methods using deep features}
We mentioned above that considering a pre-trained deep model as feature extractor to represent the images and 
train a classifier on the source provides already a strong baseline. However, we can go a step further by incorporating these representations into traditional  DA methods such as
\cite{GongCVPR12GeodesicFlowKernel,LongCVPR14TransferJointMatchingDA,FarajidavarBMVC14AdaptiveTransductiveTransfer,FernandoPRL15JointSubspaceLearningUDA,BaktashmotlaghBC17LearningDomainInvariantEmbeddingsByMatchingDistributions,CourtyPAMI17OptimalTransportDA}. As shown in
~\cite{DonahueICML14DeCAFActivationFeature,TommasiBC17ADeeperLookAtDatasetBias,SunAAAI16ReturnFrustratinglyEasyDA,CsurkaTASKCV17DiscrepancyBasedNetworksUnsupervisedDAComparativeStudy},  to cite  a few examples, using such DA methods with deep features  yields further performance improvement  on the target data.   Nevertheless,  it was observed that
the contribution of using deep features is much more significant than the contribution of using various DA methods. Indeed, as Fig. \ref{fig:BovAlexNet}) illustrates 
the gain obtained with any DA on the BOV baseline is low compared to the gain between BOV  {\em versus} deep features  both for the baseline or any DA  method.

\begin{figure}[ttt]
\begin{center}
\includegraphics[width=0.48\textwidth]{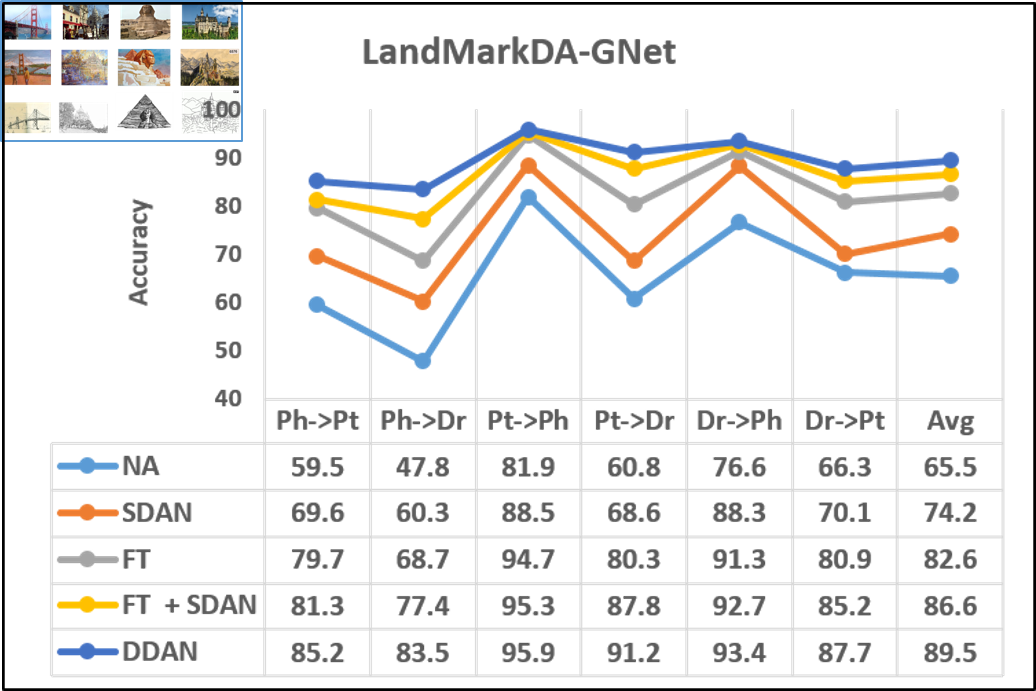}
\caption{We compare several strategies on the
LandMarkDA dataset \cite{CsurkaTASKCV17DiscrepancyBasedNetworksUnsupervisedDAComparativeStudy}
 using shallow (SDAN) and deep (DDAN) discrepancy-based networks \cite{CsurkaTASKCV17DiscrepancyBasedNetworksUnsupervisedDAComparativeStudy} built with  GoogleNet \cite{SzegedyCVPR15GoingDeeperWithConvolutions} as  backbone. No adaptation (NA) means that only the classifier layer was trained, contrary to fine-tuning the  model on the source (FT). SDAN is trained with deep features from the ImageNet pre-trained network (SDAN) or from the  fine-tuned network (FT+SDAN). We can see that FT+SDAN yields results close to DDAN, which performs the best.}
\label{fig:DeepStrategies}
\vspace{-0.4cm}
\end{center}
\end{figure}

\myparagraph{Training deep architectures on the source}
The second  solution is to train or fine-tune a deep network on the source domain  and use directly the model to
predict  the class labels for the target instances. While, in this case there is no adaptation to the target,  as 
illustrated also in Fig. \ref{fig:DeepStrategies},  we observe not only better performance (or equally if ImageNet is the source) compared with the baseline (classifier trained with the features from  backbone pretrained on ImageNet),  but also with the previous strategy (shallow DA applied with the corresponding image representations). The explanation is that  the deep model disregards in certain measure the appearance variation by focusing on high level semantics, and therefore is able to overcome in certain measure the domain gap. However,  if the domain difference between the source and target is important, fine-tuning the model on the source can also overfit the model for the source \cite{ChopraICMLWS13DLIDInterpolatingBetweenDomains,SunAAAI16ReturnFrustratinglyEasyDA} and therefore it is important  to correctly select the layers to be fine-tuned  \cite{ChuTASKCV16BestPracticesForFineTuning,CsurkaTASKCV17DiscrepancyBasedNetworksUnsupervisedDAComparativeStudy}.

\begin{figure*}[ttt]
\begin{center}
\includegraphics[width=0.92\textwidth]{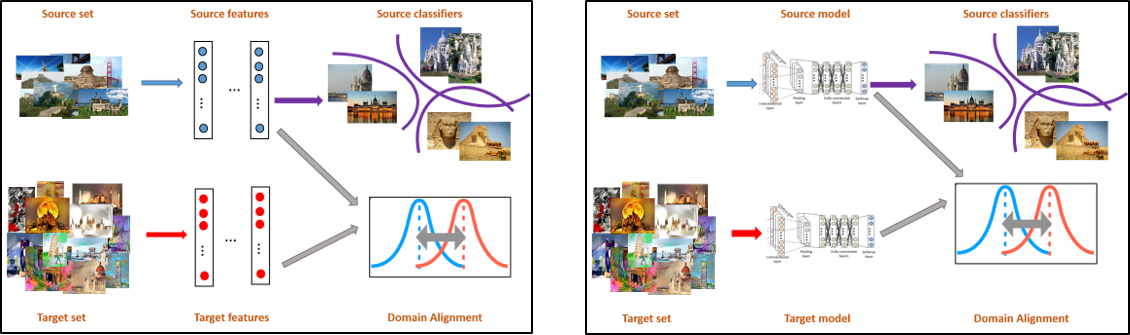}
\caption{Left: classical DA methods  where the image representations are fixed and the domain alignment and source classifier are learned in this feature space.  Right: deep DA architecture  where image representations,  source classifier and  domain alignment are all learned jointly in an end-to-end manner. The parameters of the source and target models can be partially or fully shared. }
\label{fig:ShallowvsDeeps}
\vspace{-0.4cm}
\end{center}
\end{figure*}

\myparagraph{Shallow methods using fine-tuned deep features}
Note that the above mentioned  two  strategies are orthogonal and they can be combined to take advantage of both. This is done by first fine-tuning the model on the source set and then the features extracted with this model 
are used by the shallow DA method to
decrease the discrepancy between  source and target distributions. In addition to further boosting the performance (see Fig. \ref{fig:DeepStrategies}), further advantages of this strategy are the fact that it does not require tailoring the network architecture for DA, and the fine-tuning on the source can be done in advance, even before seeing the target set.

In Fig. \ref{fig:DeepStrategies} we compare these strategies with a corresponding shallow (single layer perceptron on top of the pre-extracted features) and a deep end-to-end architecture where we use the same discrepancy (kernelized MMD ~\cite{BorgwardtBI06IntegratingKernelMaximumMeanDiscrepancy,LongICML15LearningTransferableFeaturesDAN}
and cross-entropy loss. We can see that using a shallow method with deep features extracted from the fine-tuned model indeed combines the advantages of the fine-tuning with domain adaptation and yields results close to the deep Siamese discriminative network designed for the domain adaptation. Similar behaviour was observed in  when comparing  DeepCORAL \cite{SunTASKCV16DeepCORALCorrelationAlignment} with CORAL \cite{SunAAAI16ReturnFrustratinglyEasyDA} using features extracted from the pre-trained  and fine-tuned network. Note nevertheless that in both cases a relatively simple deep DA method was  considered, and  as will be  discussed in the next sections, these deep models can be further improved in various ways.

\section{Deep DA Models}

Historical shallow DA methods include data re-weighting, metric learning, subspace representations or  distribution matching
(see for more details the surveys \cite{GopalanB15DomainAdaptationVisualRecognition,CsurkaBC17AComprehensiveSurveyDAForVisualApplications}). As discussed above, these methods  assume that the  image representations are fixed (they are handcrafted or pre-extracted from a deep model) and the adaptation model uses these features as input (see left image in Fig.~\ref{fig:ShallowvsDeeps}). 
Amongst the most popular shallow DA approaches, a set of  methods focuses on aligning the marginal distributions  of the  source and the target sets.  These methods  learn either a linear projection or more complex feature transformations with the aim that in the new  space the discrepancy between the domains is significantly decreased. Then the classifier trained on the labeled source set in the projected space, thanks to the domain alignment,  can directly be applied to the target set.
 
It is  therefore not surprising that amongst the  first deep DA models  we find the generalization of this pipeline, as illustrated in Fig.~\ref{fig:ShallowvsDeeps}(right) where the deep representation is jointly learned  with the source classifier and domain alignment in an end-to-end manner.  
These first solutions were 
followed by a large amount of different deep DA methods and architectures that can be grouped together according to different criterion (see also \cite{WangNC18DeepVisualDASurvey}). In which follows, we recall some of the main trends.

 \myparagraph{Discriminative models}
These models, inspired by classical DA methods, have a Siamese architecture~\cite{Bromley93IJPRAISignatureVerificationTimeDelayNN} with  two streams, one for the source set and one for the target set.  The two streams can share entirely, partially or not at all the weights, and in general both branches are initialized by the corresponding backbone  (\eg VGG \cite{SimonyanX14VeryDeepConvolutionalNetworks}, 
ResNet \cite{HeCVPR16DeepResidualLearning} or GoogleNet \cite{SzegedyCVPR15GoingDeeperWithConvolutions}), trained on the source set most often using the cross-entropy classification loss. 
The Siamese network is then trained with the same cross-entropy loss applied only the source stream together with a domain alignment loss defined with both source and target features. This loss uses either the last activation layer before the soft-max prediction
\cite{GhifaryPRICAI14DomainAdaptiveNN} or it can be applied to several activation layers \cite{LongICML15LearningTransferableFeaturesDAN}.

\begin{figure*}[ttt]
\begin{center}
\includegraphics[width=0.98\textwidth]{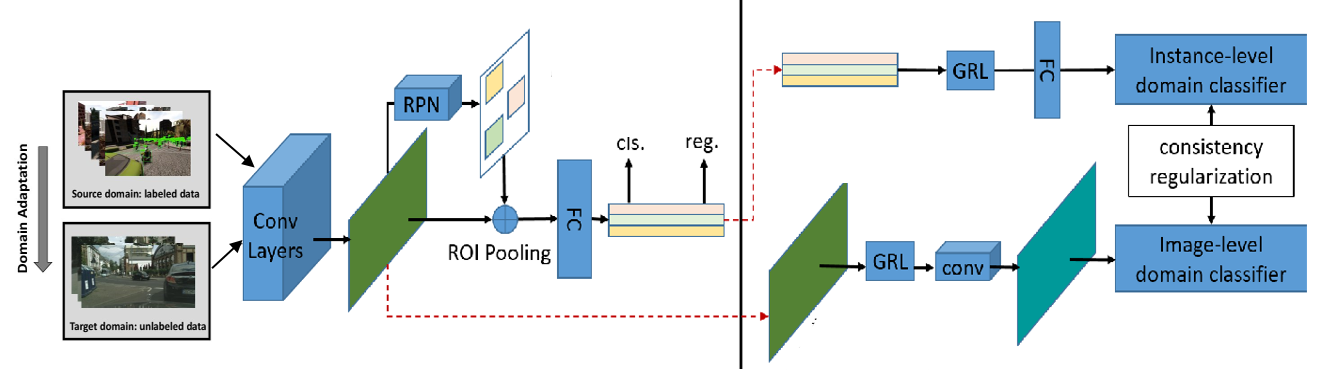}
\caption{Domain Adaptive Faster R-CNN model \cite{ChenCVPR18FasterRCNNObjDetWild} aiming to adapt the detector trained on the  source for a new domain. 
The domain shift is tackled in an adversarial training manner with GRL  \cite{GaninJMLR16DomainAdversarialNN} layers  on two levels, the image level and the instance level.  A consistency regularizer is incorporated
within these two classifiers to learn a domain-invariant region proposal network (RPN). (Image Courtesy to Yuhua Chen).}
\label{fig:objdet}
\vspace{-0.4cm}
\end{center}
\end{figure*}

The domain alignment can be  achieved by 
minimizing the feature distribution discrepancy,  or by using an adversarial loss to increase domain confusion.
To minimize the distribution discrepancy, 
most often the Kernelized MMD loss is used \cite{GhifaryPRICAI14DomainAdaptiveNN,LongICML15LearningTransferableFeaturesDAN}, but amongst the alternative losses  proposed, we can mention the Central Moment Discrepancy \cite{ZellingerICLR17CentralMomentDiscrepancyDA},
CORAL loss \cite{SunTASKCV16DeepCORALCorrelationAlignment}, 
or Wasserstein distance \cite{ShenAAAI18WassersteinDistanceGuidedRepresentationLearningDA,BalajiICCV19NormalizedWassersteinDistanceAdvLearnDA}.
Note that the Wasserstein distance is used also to minimize
the global transportation cost in optimal transport based DA methods \cite{CourtyPAMI17OptimalTransportDA,DamodaranECCV18DeepJDOTOptimalTransportUDA,XuCVPR20ReliableWeightedOptimalTransportUDA},  however, these are asymmetric models transporting the source data towards the target samples instead of projecting both sets into a 
common latent space.

On the other hand, domain confusion can be achieved either with adversarial losses such as GAN loss 
~\cite{GoodfellowNIPS14GenerativeAdversarialNets,TzengCVPR17AdversarialDiscriminativeADDA,VolpiCVPR18AdversarialFeatureAugmentationUDA} and domain confusion loss \cite{TzengICCV15SimultaneousDeepTL,GebruICCV17FineGrainedRecognitionWildMultiTaskDA}, or by using a domain classifier and gradient reversal layer (GRL)~\cite{GaninJMLR16DomainAdversarialNN,PeiAAAI18MultiAdversarialDA}.  Note however that the latter can also be formulated as a min-max loss and is achieved by 
the integration of a simple binary domain classifier  
and a GRL layer into a standard deep architecture which 
is unchanged during  the forward pass, and reversed for the target during backpropagation. This simple but 
quite  powerful solution  became extremely popular when DA is applied for problems beyond image classification, in particular for object detection \cite{ChenCVPR18FasterRCNNObjDetWild,SaitoCVPR19StrongWeakDistributionAlignmentObjDet,ZhuCVPR19AdaptingObjDetSelectiveCrossDomainAlignment,HeICCV19MultiAdversarialFasterRCNNUnrestrictedObjDet,XuCVPR20ExploringCategoricalRegularizationDAObjDet} (see also Fig. \ref{fig:objdet}), semantic image segmentation \cite{HoffmanX16FCNsInTheWildPixelLevelAdversarialDA,TsaiCVPR18LearningToAdaptStructuredOutputSpaceSemSegm} or video action recognition \cite{LiNIPS18UnsupervisedLearningViewInvariantActionRepr,MunroICCVWS19MultiModalDAFineGrainedActionRecognition}.

\myparagraph{Class-conditional distribution alignment}
To overcome the drawback that aligning marginal distributions without taking   into account explicitly the task might lead to sub-optimal solution, several approaches were proposed. 
Amongst them we have the ones that tries to  align class conditional distributions  by  minimizing the  marginals of features and class predictions jointly \cite{LongICML17DeepTLJointAdaptationNetworks}, or  exploit discriminative information conveyed in the classifier predictions to assist adversarial adaptation 
 \cite{LongNIPS18ConditionalAdversarialDomainAdaptation}.
Instead,  \cite{ZhangICML19BridgingTheoryAlgorithmDA} 
proposes  to focus on the  Margin Disparity Discrepancy  loss defined on the scoring function and use adversarial learning  to solve it. 
\cite{SaitoCVPR18MaximumClassifierDiscrepancyUDA, SaitoICLR18AdversarialDropoutRegularization} proposes  to minimize task-specific decision boundaries’ disagreement on target examples while aligning features across domains.  \cite{KangCVPR19ContrastiveAdaptationNetworkUDA} explicitly models the intra-class  and the inter-class domain discrepancy,  where  intra-class domain discrepancy is minimized to avoid  misalignment  and  the inter-class domain discrepancy is
maximized to enhance the model’s generalization ability. 
Assuming the access to at least a  small set of labeled target samples, \cite{KoniuszCVPR17DomainAdaptationMixtureAlignmentsSecondHigherOrderScatterTensors}  proposed to align higher-order scatter statistics between 
domain-specific and class-specific representations.

\myparagraph{Network parameter adaptation} The above methods in general keep the same architecture with the same weights for both source and target streams, which essentially aims to learn domain invariant features. In
contrast to them, several  approaches were proposed, where  the  goal is to specialize the streams 
for the respective domains by adapting the parameters of the target stream. As such, \cite{RozantsevPAMI18BeyondSharingWeightsDeepDA,RozantsevCVPR18ResidualParameterTransferDeepDA} explicitly model the domain shift by learning meta parameters that transform the weights and biases of each layer of the network from the source stream to the target one. 
Instead, \cite{BermudezChaconICLR20DomainAdaptiveMultibranchNetworks}  consider a multi-stream architectures with non shared parameters where learnable gates at multiple levels
allows the network to find for each domain a corresponding weighted aggregation of these parallel streams.

\begin{figure*}[ttt]
\begin{center}
\includegraphics[width=0.9\textwidth]{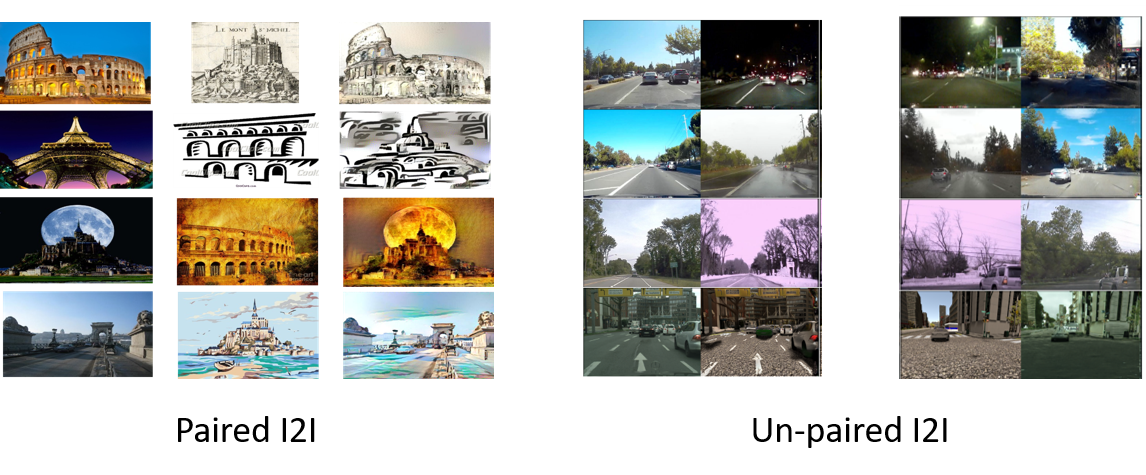}
\caption{Left: Paired image style transfer \cite{GatysNIPS15TextureSynthesisCNN}  where the model takes the content of the source images (first column) and the style of the target image (second column) to generate a target-like source image (third column). Note that these images inherits the label from the source while they look more like the target images.
Right:  Un-paired image-to-image (I2I) transfer where the model learns to synthesize directly target-like images (night, rainy, {\em etc}) for a source input and/or source-like images (day, sunny, {\em etc}) for a target image without the need of an explicit style image.}
\label{fig:styleTransfer}
\vspace{-0.4cm}
\end{center}
\end{figure*}

\myparagraph{Domain specific batch normalization} 
\cite{CarlucciICCV17AutoDIALDomainAlignmentLayers,LiPR18AdaptiveBatchNormalizationForPracticalDomainAdaptation,ChangCVPR19DomainSpecificBatchNormalizationUDA} have shown that  domain specific batch normalization  is equivalent to projecting the source and target feature distributions to a reference distribution through feature standardization.  
Hence this yields a simple yet efficient solution for 
minimizing the  gap between  domains. \cite{CuiCVPR20TowardsDiscriminabilityDiversityBatchNuclearNormMaximization} proposes 
batch nuclear-norm maximization to simultaneously enhance the discriminability and diversity of predicted scores. \cite{ManciniCVPR19UnifyingPredictiveContinuousDAthroughGraphs} applied domain-specific batch normalization
layers in the context of graph-based predictive DA. 
\cite{PerrettCVPR19DDLSTMDualDomainLSTMActionRecogn} proposes the DDLSTM architecture for action recognition that  performs cross-contaminated recurrent batch normalisation for   both single-layer and multi-layer LSTM architectures.

\myparagraph{Encoder–decoder reconstruction}
Early  deep auto-encoder frameworks proposed for DA in  NLP 
 \cite{GlorotICML11DASentimentClassification}
rely on the feedforward stacked denoising
autoencoders \cite{VincentICML08ExtractingDenoisingAutoencoders} where a multi-layer  neural network  reconstructs the input data from partial random corruptions  with backpropagation.  \cite{ChenICML12MarginalizedDenoisingAutoencodersDA}  has shown that such model can be trained efficiently by marginalizing out the noise that leads  to a closed form  solution for the transformations  between layers. 
\cite{CsurkaTASKCV16UnsupervisedDAInstanceDenoising} extended this unsupervised network to a supervised one by jointly learning the domain invariance  with the cross-domain classifier while keeping the network solvable in a single forward pass. 

In contrast to these models that act on the pre-extracted features, more recent reconstruction models trains the encoders/decoders end-to-end.  As such,  \cite{GhifaryECCV16DeepReconstructionClassificationNetworkDA} combines the standard CNN for source label prediction
with a deconvolutional network~\cite{ZeilerCVPR10DeconvolutionalNetworks} for target data reconstruction  by alternating  between
unsupervised and supervised  training. 
 \cite{BousmalisNIPS16DomainSeparationNetworks}  integrates both domain-specific  encoders and  shared encoders,  and the model integrates a reconstruction loss for a shared decoder that rely on both domain specific  and shared representations.

\myparagraph{Transfer domain  style} 
In many cases the domain shift between domains is strongly related to the image appearance change such as day to night, seasonal change, synthetic to real. Even stronger domain shift can be observed when the adaptation is aimed to be between  images that exhibit different  artistic style such as paintings, cartoons and sketches~\cite{CastrejonCVPR16LearningAlignedCrossModal,LiCVPR17DeeperBroaderArtierDG,CsurkaTASKCV17DiscrepancyBasedNetworksUnsupervisedDAComparativeStudy}.  To explicitly
account for such  stylistic domain shifts, a set of papers  proposed to use  image-to-image (I2I) style transfer  methods \cite{GatysNIPS15TextureSynthesisCNN,HuangICCV17ArbitraryStyleTransferRealTimeAdaptiveInstanceNormalization,LiECCV18ClosedFormImageStylization} to generate a  set of  {\em target like} source images. They  have shown that this new set is suitable to train a model for the target set \cite{CsurkaTASKCV17DiscrepancyBasedNetworksUnsupervisedDAComparativeStudy,ThomasACCV19ArtisticObjectRecognitionUnsupervisedStyleAdaptation}. The main reason why this works is that  these synthesized images  inherits
the semantic content of the source, and hence its label, 
while their appearances  is more similar to the target style (see examples in Figure \ref{fig:styleTransfer}(Left)).
Training a model with this set not only outperforms  the model trained with the original source set,  but it is also easier to further adapt it to the target set \cite{CsurkaTASKCV17DiscrepancyBasedNetworksUnsupervisedDAComparativeStudy}. 

\begin{figure*}[ttt]
\begin{center}
\includegraphics[width=0.9\textwidth]{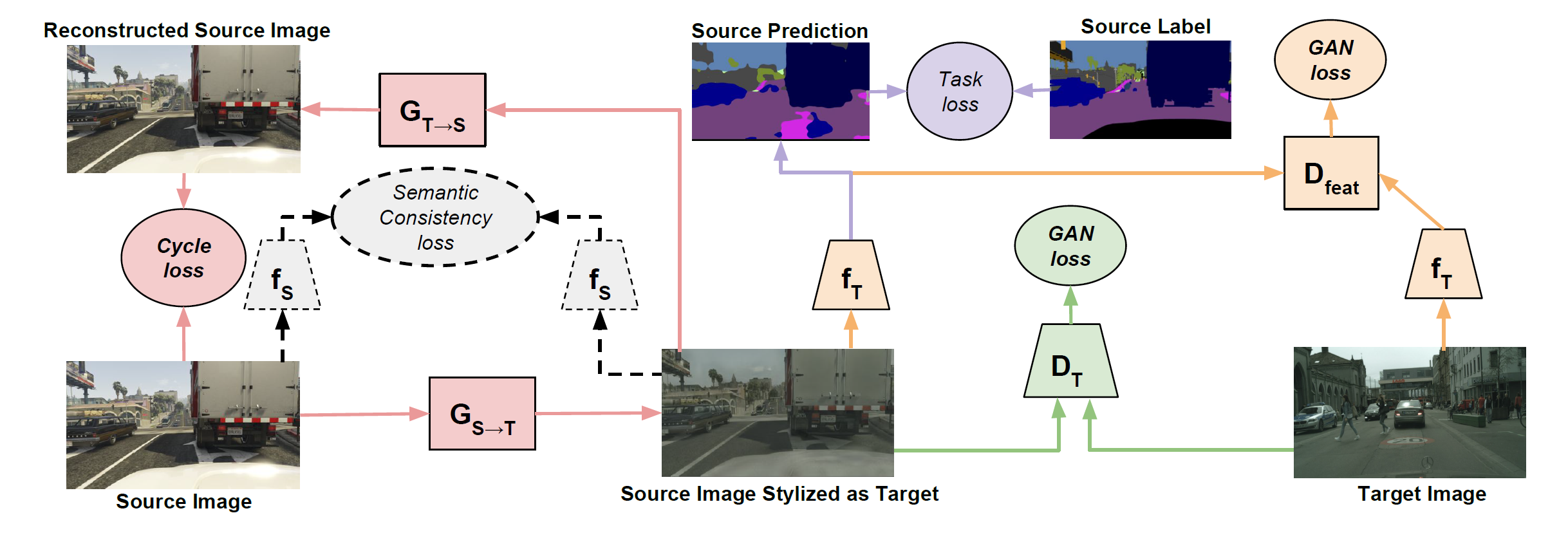}
\caption{CyCADA \cite{HoffmanICML18CyCADACycleConsistentAdversarialDA}, combines  pixel-level and feature-level adaptation where both structural and semantic consistency is enforced. The former is ensured by an L1 penalty on the reconstruction
error between the source image and the image reconstructed from the target-like source. To ensure the latter, a semantic consistency loss is used that forces the segmentation of the target-like source image to match 
the  source predictions. (Image Courtesy to Judy Hoffman).}
\label{fig:cycada}
\vspace{-0.4cm}
\end{center}
\end{figure*}

Another set of methods seek to  learn how to translate
between domains without using paired input-output examples but instead assuming there is some underlying appearance shift between the domains (\eg day to night, sunny to rainy, synthetic to real). For example, \cite{YooECCV16PixelLevelDomainTransfer,BousmalisCVPR17UnsupervisedPixelLevelDAGAN,TaigmanICLR17UnsupervisedCrossDomainImageGeneration} train the network 
to synthesize target-like and/or source-like images (see Figure \ref{fig:styleTransfer}(Right)) in general by relying on a Generative Adversarial Networks (GANs) \cite{GoodfellowNIPS14GenerativeAdversarialNets}, 
where an adversarial loss force the model to
generating fake (target-like) images to be indistinguishable from real (target) photos. A pair of GANs,  each   corresponding to  one of the domains is considered in \cite{LiuNIPS16CoupledGenerativeAdversarialNetworks}, where the  model adapts the input noise vector to paired images that are from the two distributions and share the labels.
This work was extended in 
\cite{LiuNIPS17UnsupervisedI2ITranslationNetworks}  with Variational Auto-Encoders (VAE), where 
the  image reconstruction, image translation, and the cycle-reconstruction are jointly optimized. 
\cite{ZhuICCV17UnpairedI2ICycleConsistentAdversarialNetworks} proposes to learn a mapping between  source and  target domains  using an adversarial GAN loss while imposing a cycle consistent  loss, \ie  the target-like source image  mapped  back to source style  should match the original source image. 
 \cite{HoffmanICML18CyCADACycleConsistentAdversarialDA} 
combined cycle consistency between 
input and stylized images with task-specific semantic 
consistency, and extended the method to semantic segmentation (see Figure \ref{fig:cycada}). Transferring the target image style to generate synthetic source images is at the  core of many DA method for semantic segmentation
\cite{MurezCVPR18ImageToImageTranslationDA,SankaranarayananCVPR18LearningSyntheticDataDomainShiftSemSegm,WuECCV18DCANDualChannelWiseAlignmentNetworksUDA,ChangCVPR19AllAboutStructureDASemSeg,YangCVPR20PhaseConsistentEcologicalDA}. GAN-like DA models combined with similarity preserving constraints were often used  for adapting
cross-domain person re-identification models \cite{BakECCV18DomainUDAASynthesisReId,DengCVPR18ImageImageDAwithPreservedSelfSimilarityReID,ChenICCV19InstanceGuidedContextRenderingDAReID}.

\section{Orthogonal improvement strategies}

In addition to the specifically tailored deep DA architectures, several 
machine learning strategies can be used with the above models to further improve their performance. While, in some cases such methods were used the main DA solution,  we discuss them here separately,  as in general these ideas can be easily combined with most of the above mentioned DA models.

\myparagraph{Pseudo-labeling the target data} One of the most used such technique is  self-supervised learning   with pseudo-labeled target data, sometimes referred to as self-labeling or self-training. The underlying assumption here is that at least for a subset of target samples the labeling is correct and hence the model can  rely on them  to improve the model. 
In this way the model acts as if it was a semi-supervised DA model, except that instead of having ground-truth target labels, these labels come from a pseudo-labeling process. 
As not all predictions are correct, often pseudo-labeling confidence scores are computed and used to select  which pseudo-labeled  samples should be retained for training. 
Typical approaches to obtain  pseudo labels are, using
the softmax predictions  \cite{SaitoICML17AsymmetricTriTrainingUDA,DengTCSVT20RethinkingTripletLossDA}, using distance to class prototypes \cite{CsurkaTASKCV14DomainDomainSpecificClassMeans,PanCVPR19TransferrablePrototypicalNetworksUDA},
clustering \cite{KangCVPR19ContrastiveAdaptationNetworkUDA,SharmaWACV20UnsupervisedMetaDAFashionRetrieval},
label propagation on the joint source-target nearest neighbour graph \cite{TommasiICCV13FrustratinglyEasyDA,SenerNIPS16LearningTransferrableRepresentationsUDA}, 
via augmented anchors
\cite{ZhangECCV20LabelPropagationAugmentedAnchorsUDA},  or even
considering  a teacher classifier, built as an implicit ensemble of  source classifiers \cite{DengICCV19ClusterAlignmentTeacherUDA}.
 
Self-supervising deep DA models with pseudo-labeled target samples is also a popular  strategy used to adapt 
tasks beyond image classification.  For example,  
\cite{SharmaWACV20UnsupervisedMetaDAFashionRetrieval} proposed several  strategies to pseudo-label fashion products across datasets and use them to solve the meta-domain gap  occurring between consumer and shop fashion images. 
\cite{GeX20StructuredDAOnlineRelationRegularizationReID} proposed a DA framework with online relation regularization for person re-identification that uses target pseudo labels to improve the target-domain encoder trained  via a joint cross-domain labeling system. \cite{LiCVPR19BidirectionalLearningDASemSegm} used predicted labels with high confidence in a 
 bidirectional learning framework for semantic segmentation, where the image translation model and the segmentation adaptation model are learned alternatively. \cite{WangCVPR20DifferentialTreatmentStuffThingsDASemSegm} combines the self-supervised learning strategy
with a framework where the model is disentangled into a "things" and a "stuffs" segmentation networks.

\myparagraph{Curriculum learning}
To minimise the impact of noisy pseudo-labels during
alignment, curriculum learning-based  \cite{BengioACMMM09CurriculumLearning}  approaches have
been explored. A simple and most used  curriculum learning scenario in DA is to  first consider the
most confident target samples for the alignment and including the less confident ones at  later stages of the training. Pseudo-labeling confidence
scores are typically determined using the image classifiers
\cite{RoyCVPR19UDAFeatureWhiteningConsensusLoss,ZhangCVPR18CollaborativeAdversarialUDA}, similarity to neighbours \cite{SenerNIPS16LearningTransferrableRepresentationsUDA,TommasiICCV13FrustratinglyEasyDA} or to class
prototypes \cite{ChenCVPR19ProgressiveFeatureAlignmentUDA, CsurkaTASKCV14DomainDomainSpecificClassMeans}. After each epoch, \cite{ZhangCVPR18CollaborativeAdversarialUDA} increases the training set with new target samples that are both highly confident and domain uninformative. To improve the confidence of pseudo-labels, \cite{RoyCVPR19UDAFeatureWhiteningConsensusLoss} relies on the consensus of image transformations, whereas  \cite{SaitoICML17AsymmetricTriTrainingUDA} considers the agreement between multiple classifiers. \cite{ShuAAAI19TransferableCurriculudmWeaklySupervisedDA} proposes a   weakly-supervised DA framework that
 alternates  between quantifying the transferability of source examples based on their contributions to the target task and   progressively integrating from easy to
hard examples.
\cite{KangCVPR19ContrastiveAdaptationNetworkUDA} considers  target clusters  initialized  by the  source cluster centers, and assign target samples to them.  At each epoch, first target elements  that  are far from the affiliated cluster are discarded,  then the clusters with too few target samples assigned are also discarded.

Curriculum-learning based DA methods with  progressively including harder and harder pseudo-labeled target data  was also used  for cross-domain person re-identification \cite{FanX17UnsupervisedPersonReIDClusteringFineTuning,ZhangICCV19SelfTrainingWithProgressiveAugmentation,FuICCV19SelfSimilarityGroupingDAPersonReID} and  image segmentation \cite{ZouECCV18UnsupervisedDASemSegmClassBalancedSelfTraining,DuICCV19SSFDANSeparatedSemanticFeatureDASemSegm,PanCVPR20UnsupervisedIntraDASemSegmSelfSupervision}.

\myparagraph{Conditional entropy minimization} 
Widely used to  improve the performance of semi-supervised learning, conditional entropy minimization 
in the target domain is another way to improve 
decision boundaries of the model \cite{CarlucciICCV17AutoDIALDomainAlignmentLayers,SaitoICML17AsymmetricTriTrainingUDA,ShuICLR18ADIRTTApproachUDA,LongNIPS18ConditionalAdversarialDomainAdaptation}.
The Minimax Entropy  loss
\cite{SaitoICCV19SemiSupervisedDAMinimaxEntropy}  is a variant where an adversarial learning  maximizes the conditional entropy of unlabeled target data with respect to the classifier and minimizes it with respect to the feature encoder. Similarly, \cite{VuCVPR19ADVENTAdversarialEntropyMinimizationDASemSegm} proposes an adversarial loss for  entropy minimization used to bridge the domain gap between synthetic to real semantic segmentation adaptation. 
\cite{RoyCVPR19UDAFeatureWhiteningConsensusLoss} proposes the Min-Entropy Consensus that merges both the entropy and the consistency loss into a single unified function.

\myparagraph{Self-ensemble learning}
The main idea of self-ensemble learning is 
to train the neural network  with small perturbations 
such as different augmentations, using dropout and various noise  while forcing the network 
to make consistent predictions for the target samples. 
In this spirit,
\cite{KurmiBMVC19CurriculumBasedDropoutDiscriminatorDA}, proposed a Monte Carlo dropout based ensemble discriminator
by gradually increasing the variance of
the sample based distribution. \cite{FrenchICLR18SelfEnsemblingForVisualDA} extended the idea of learning with a mean teacher network \cite{TarvainenNIPS17MeanTeachersBetterRoleModelsSSL} to domain adaptation considering  a separate
path for source and target sets and sampling  independent batches  making the batch normalization domain specific during  the training process.
\cite{DengICCV19ClusterAlignmentTeacherUDA}
builds a teacher classifier, to provide pseudo-labels used by a class-conditional clustering loss to force the features from the same class to concentrate together and a conditional feature  matching loss to align the clusters from different domains.


\end{document}